\documentclass[11pt,a4paper]{article}
\usepackage[hyperref]{acl2021}
\usepackage{times}
\usepackage{latexsym}

\usepackage{microtype}

\aclfinalcopy 

\usepackage{booktabs}
\usepackage{makecell} 
\usepackage{multirow}
\usepackage{multicol}
\usepackage[ruled]{algorithm2e}
\DontPrintSemicolon
\usepackage{adjustbox}
\usepackage{float}
\usepackage{caption}
\usepackage{subcaption}
\usepackage{graphics}
\newsavebox{\tempbox}
\usepackage{amsmath,amsthm,amssymb}
\usepackage{mathtools}
\usepackage{xspace}

\makeatletter
\newcommand{\mypm}{\mathbin{\mathpalette\@mypm\relax}}
\newcommand{\@mypm}[2]{\ooalign{%
  \raisebox{.1\height}{$#1+$}\cr
  \smash{\raisebox{-.6\height}{$#1-$}}\cr}}
\makeatother

\DeclarePairedDelimiterX{\infdivx}[2]{(}{)}{%
  #1\;\delimsize|\delimsize|\;#2%
}
\newcommand{\kld}[2]{\ensuremath{D_{KL}\infdivx{#1}{#2}}\xspace}

\newcommand{\bo}[1]{\boldsymbol{#1}}

\makeatletter
\newcommand{\removelatexerror}{\let\@latex@error\@gobble}
\makeatother

\title{Towards Zero-Shot Knowledge Distillation for Natural Language Processing}

\author{
 Ahmad Rashid\textsuperscript{1}, Vasileios Lioutas\textsuperscript{2}\thanks{$\;\;$Work done during an internship at Huawei Noah’s Ark Lab.} , Abbas Ghaddar\textsuperscript{1}, Mehdi Rezagholizadeh\textsuperscript{1} \\
 \textsuperscript{1}Huawei Noah’s Ark Lab, \textsuperscript{2}University of British Columbia\\
 \texttt{\normalsize ahmad.rashid@huawei.com,}
 \texttt{\normalsize contact@vlioutas.com,} \\ \texttt{\normalsize abbas.ghaddar@huawei.com,}  \texttt{\normalsize mehdi.rezagholizadeh@huawei.com}
}

\date{}

\begin{document}
\maketitle
\begin{abstract}
Knowledge Distillation (KD) is a common knowledge transfer algorithm used for model compression across a variety of deep learning based natural language processing (NLP) solutions. In its regular manifestations, KD requires access to the teacher's training data for knowledge transfer to the student network. However, privacy concerns, data regulations and proprietary reasons may prevent access to such data. We present, to the best of our knowledge, the first work on Zero-Shot Knowledge Distillation for NLP, where the student learns from the much larger teacher without any task specific data. Our solution combines out of domain data and adversarial training to learn the teacher's output distribution. We investigate six tasks from the GLUE benchmark and demonstrate that we can achieve between 75\% and 92\% of the teacher's classification score (accuracy or F1) while compressing the model 30 times.
\end{abstract}

\section{Introduction}

Deep Learning based Natural Learning Processing (NLP) systems have become state-of-the-art on many applications such as Machine Translation (MT)~\citep{vaswani2017attention,time-awareconv}, Natural Language Understanding (NLU)~\citep{devlin-etal-2019-bert} and Language Generation~\citep{brown2020language} among others. These models are increasingly trained on huge corpora and with billions of trainable parameters~\citep{brown2020language}. This is prohibitive for deploying these models on edge devices as well as maintaining them on servers. Moreover, training and evaluating them leaves a significant environmental footprint~\citep{strubell2019energy}  wherein avoiding the resource hungry training is very challenging and may be unavoidable~\citep{li2020train}. Model compression approaches make it feasible to employ current state of the art models on edge devices.

Model Compression~\citep{sanh2019distilbert,jiao2019tinybert} has received a lot of attention in the NLP community due to the aforementioned reasons. Some of the algorithms include model pruning~\citep{see2016compression}, quantization~\citep{shen2019q}, low-rank matrix factorization~\citep{sainath2013low} and knowledge distillation (KD)~\citep{bucilua2006model,hinton2015distilling}. 

KD is one of the most commonly used, application and model agnostic, compression and ensembling algorithm. Some of the advantages of KD include use of just the teacher logits and architecture independence between the student and the teacher. However, the student needs to be trained with the teacher's training data so as to prevent loss of accuracy. We can not assume this access for many practical problems. Some of the concerns preventing access include data privacy, intellectual property, size and transience~\citep{micaelli2019zero}. e.g. a model trained on patient health records might be available but the data itself may be inaccessible due to patient privacy.

In computer vision (CV), Zero-Shot KD (ZSKD) has been proposed to train a student without using any data. In this context Zero-Shot refers to training without using data instead of no training at all. \citet{nayak2019zero} propose generating "data impressions" by updating noise using backpropogation until it generates  'valid' teacher logits and then training the student on these data impressions. \citet{chen2019data} use a generator to produce synthetic images and use the teacher as discriminator, observing that for real images the softmax function of the teacher encourages a unimodal distribution. \citet{micaelli2019zero} use a generator to produce synthetic training samples employing adversarial training to improve the quality. \citet{yoo2019knowledge} generate synthetic data by conditioning a generator on output samples from the teacher and a low dimensional representation of the generated samples. These works assume that there is no data available whatsoever for training the student. However, we contend that a similar approach for NLP does not work. We relax this condition and argue that we can still achieve the goals of ZSKD if we use easy to access out-of-domain (OOD), task agnostic data to aid the process. \citet{krishna2019thieves} put forth a similar argument, albeit for the problem of model extraction, where they use simple heuristic rules to generate training data for a student, of similar or larger size to the teacher, in order to learn the teacher's output distribution. However, they do not put constraints on the size of the student and even propose a student larger than the teacher. Moreover they assume only API access to the teacher with a fix budget. 

We study the problem of ZSKD for NLP and in particular present the following contributions:

\begin{itemize}
    \item We present one of the first works in NLP on model compression for NLU models using KD without the teacher's training data or any other task-specific data.
    \item We present a novel KD algorithm which combines OOD data and adversarial training.
    \item Our algorithm generalizes to different classification tasks for NLP including sentiment analysis, question answering, entailment etc.
    \item We present an analysis of our algorithm on Natural Language Inference. 
\end{itemize}

\section{Related Work}
\subsection*{Knowledge Distillation} 
KD \citep{hinton2015distilling} is a well-known deep learning technique to transfer the knowledge from an already trained large teacher model to a smaller student network. KD adds a new loss function to the student's regular training loss over the training labels. This new loss function aims at matching the smoothened output probabilities of the student with those of the teacher. More specifically,  the training data is fed into the teacher model and the teacher logits are obtained. These are fed, typically, into a softmax function and the temperature parameter is adjusted to smoothen the resulting label distribution. The training loss function for the KD algorithm is as following:

\begin{align}
\begin{split}
    \mathcal{L}_{KD} = & \; \alpha * \mathcal{H}(y,\sigma(z_s;T=1)) \; + \\ & \; (1-\alpha)*\mathcal{H}(\sigma(z_t;T=\tau),\sigma(z_s,T=\tau))
\end{split}
\end{align}

where $\mathcal{H}$ can be the cross-entropy or any other valid loss function and $z_s$ and $z_t$ are the student and teacher logits respectively. We will use the  Kullback–Leibler (KL) divergence in our algorithm between the teacher and student logits. $\sigma$ is the softmax function, and  $\tau$ and $\alpha$ are training parameters.  

\begin{figure*}[t]
\centering
    \includegraphics[width=1\textwidth]{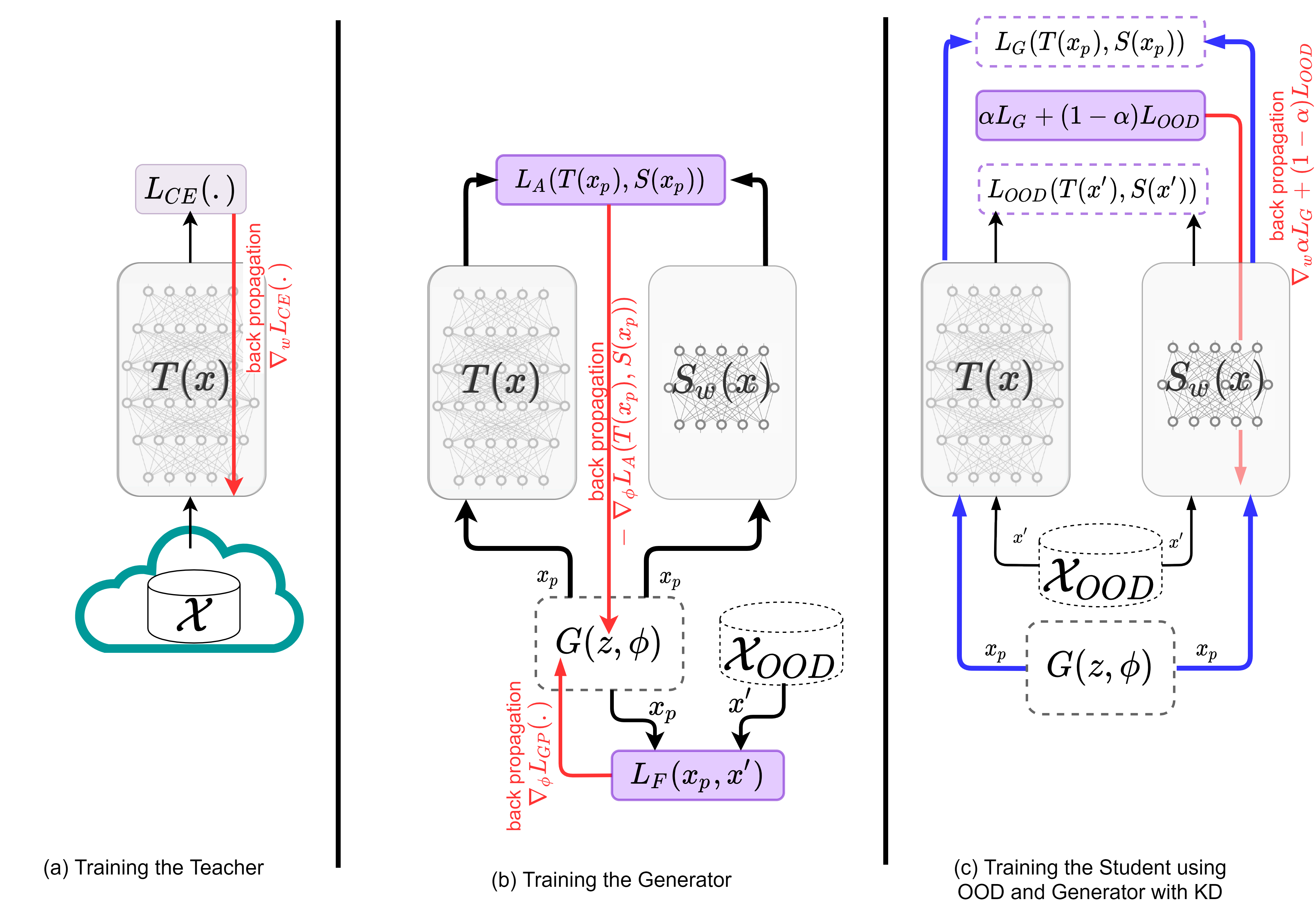}
    \caption{Schematic Diagram of our Zero-Shot KD solution. a) We assume access to a pre-trained teacher. b) We adversarially train the generator using out-of-domain data (OOD). c) Finally we use the generated data and the OOD data for KD.}

\label{fig:1}
\end{figure*}

\subsection*{Few Sample and Zero-Shot Knowledge Distillation} 

Considering that the KD training procedure is data hungry and requires access to a large training set, the Few Sample Knowledge Distillation (FSKD) technique~\cite{li2018few} deals with the sample efficiency and training efficiency of KD. The FSDK technique is comprised of three main steps which are: first, compressing the teacher network to the student network; second, aligning the student and teacher blocks by adding layers to match the size of output blocks of the student and the teacher; third, absorbing the added layers in the student network. This training process leaves only a few training parameters in the student network which makes the training possible with small amount of unlabeled samples~\cite{li2018few}. FSKD can be applied to on-device training, cloud services for customers with private data, or even fast convergence training scenarios. Our work is different from FSKD in the sense that we do not have access to any of teacher's training data.

ZSKD deals with scenarios in which either no training data is available (e.g. in ~\cite{nayak2019zero}) or at least teacher's training data is not available (for example due to customer's privacy issues).
\citet{lopes2017data} introduce a data-free knowledge distillation approach with the assumption that the teacher's network and some meta-data (i.e. the teacher activation records or statistics on the teacher's training data) are given. This work reconstructs the original training data by tweaking a noise input and trying to recover the given meta-data. We are different from~\cite{lopes2017data} in the sense that our model does not need any meta-data for training.  
Another case in point is ~\cite{nayak2019zero} which introduces a data-free knowledge distillation approach with no knowledge about the target data distribution. In this regard, their Zero-Shot technique models the softmax output of the teacher using the Dirichlet distribution and then builds the underlying data samples (so called Data Impressions ) corresponding to that modeled distribution for the teacher. This Zero-Shot technique (ZSKD) is experimented on MNIST and CIFAR-10, however, we believe that this approach will not be practical for NLP tasks due to the fact that the input data is discrete and the size of the output softmax can be really large. One potential practical scenario for NLP can be training students without accessing teacher's training data. In this scenario, we are allowed to use any text corpus in the public domain except the data used for training the teacher network. In this case we can borrow ideas from model extraction techniques such as~\cite{pal2019framework, krishna2019thieves,yoo2019knowledge} to facilitate ZSKD training by querying the teacher model using unlabeled data. \cite{pal2019framework, krishna2019thieves} deal with textual input but do not consider smaller students and the KD scenario. \cite{yoo2019knowledge} designs a conditional data generator to tackle with lack of training data for the student network and focuses on image classification. However, our solution works on text and our text generator is unconditional.   

\subsection*{Adversarial Training}
Adversarial examples are small perturbations to training samples indistinguishable to humans but enough to fool neural network classifiers. \citet{goodfellow2014explaining} proposed adding them to the training set to make CV systems robust to adversarial attacks. \citet{miyato2016adversarial} adapt adversarial training to text classification and improve performance on a few supervised and semi-supervised text classification tasks. 

Adversarial training although proposed for model robustness~\cite{ebrahimi2017hotflip}, has been shown to improve state-of-the-art model performance~\cite{cheng2019robust,zhu2019freelb} in NLP. \citet{cheng2019robust} study machine translation and propose making the model robust to both source and target perturbation, generated by swapping the word embedding of a word with that of its synonym. They model small perturbations by considering word swaps which cause the smallest increase in loss gradient. They achieve a higher BLEU score on Chinese-English and English-German translation compared to the baseline. \citet{zhu2019freelb} propose a novel adversarial training algorithm, FreeLB, to make gradient based adversarial training efficient by updating both embedding perturbations and model parameters simultaneously during the backward pass of training. They show improvements on multiple language models on the GLUE  benchmark. 

\citet{micaelli2019zero} adapt adversarial training for ZSKD and train an image generator to increase the divergence between student and teacher and train the student to decrease this divergence.


\savebox{\tempbox}{
\removelatexerror
\begin{minipage}[c]{0.46\textwidth}%
\begin{algorithm*}[H]
 \SetAlgoLined
 \SetKwInput{pretrain}{pretrain}
 \SetKwInput{data}{dataset}
 \SetKwInput{init}{initialize}
 \pretrain{$T(\cdot)$}
 \data{$D$}
 \init{$G(\cdot; \phi)$}
 \init{ $S(\cdot; \theta)$}
\BlankLine
 \BlankLine
 \For{$k \leftarrow 1, 2, ..., N$}{
    $\bo{x}_k \leftarrow D$ \;
    \BlankLine
    \# Adversarial Step \;
    \For{$1, 2, ..., n_G$}{
        $\bo{z} \leftarrow \{\bo{z_0,\ldots,z_l}\} \sim \mathcal{N}(\bo{0},\bo{std})$ \;
        $\bo{x}_\text{logits} \leftarrow G(\bo{z}; \phi)$ \;
        $\bo{x}_p \leftarrow \text{Gumbel-Softmax}(\bo{x}_\text{logits})$ \;
        $\mathcal{L}_A \leftarrow - \kld{T(\bo{x}_p)}{ S(\bo{x}_p)}$ \;
        $\mathcal{L}_F \leftarrow \kld{\bo{x}_k}{ \bo{x}_p}$ \;
        $\mathcal{L}_T \leftarrow \frac{\mathcal{L}_A + \mathcal{L}_F}{2} $ \;
        $\phi \leftarrow \phi - \eta \dfrac{\partial \mathcal{L}_T}{\partial \phi}$ \;
    }
    \BlankLine
    \BlankLine
    \BlankLine
    \BlankLine
    \BlankLine
    \BlankLine
    \BlankLine
    \BlankLine
    \BlankLine
    \BlankLine
    \# Knowledge Distillation \;
    \For{$1, 2, ..., n_S$}{
        $\bo{z} \leftarrow \{\bo{z_0,\ldots,z_l}\} \sim \mathcal{N}(\bo{0},\bo{std})$ \;
        $\bo{x}_\text{logits} \leftarrow G(\bo{z}; \phi)$ \;
        $\bo{x}_p \leftarrow \text{Gumbel-Softmax}(\bo{x}_\text{logits})$ \;
        \BlankLine
        $\mathcal{L}_G \leftarrow \kld{T(\bo{x}_p)}{ S(\bo{x}_p)}$ \;
        $\mathcal{L}_{OOD} \leftarrow \kld{T(\bo{x}_k)}{ S(\bo{x}_k)}$ \;
        $\mathcal{L} \leftarrow \alpha \cdot \mathcal{L}_G + (1-\alpha) \cdot \mathcal{L}_{OOD}$ \;
        $\theta \leftarrow \theta - \eta \dfrac{\partial \mathcal{L}}{\partial \theta}$ \;
    }
    \BlankLine
    decay $\eta$ \;
    \BlankLine
    \BlankLine
}
\end{algorithm*}
\end{minipage}}

\begin{algorithm*}[t]
\caption{Adversarial Training for Zero-Shot Knowledge Distillation}
\label{alg:distillation}
\clipbox{0pt {\depth} 0pt {\baselineskip}}{\usebox{\tempbox}}\hfill
\raisebox{\depth}{\clipbox{0pt 1ex 0pt {\height}}{\usebox{\tempbox}}}
\end{algorithm*}

\section{Methodology}

We solve the missing data problem for knowledge transfer between teacher and student. We rely on an adversarial text generator as the backbone of our method. However, we still need data to pre-train the generator. Since we assume access to this general purpose OOD data, we delineate general principles to extract a training set from this source as well. Finally, we apply KD on a combination of the OOD training data and the adversarial training data. Figure~\ref{fig:1} gives a visual illustration of the proposed ZSKD method.    

\subsection{Out-of-Domain Training Data} 
\label{sec:odd}

Our ZSKD method assumes that we do not have the original training data on which the teacher model is trained as well as any other task specific data. Similar to \cite{krishna2019thieves}, we construct an out-of-domain (OOD) dataset. The idea is that using a general purpose corpus of text, we randomly sample sentences from the text. Then depending on the task we add simple heuristics to make the text suitable for the problems at hand. We summarize a list of targeted tasks all taken from the GLUE benchmark.

\textbf{Sentiment Classification (SST-2)}. We do not modify the sampled sentences for this task but simply feed them to the teacher to get the sentiment output distribution, even though most sentences in the sampled text would have neutral sentiment.

\textbf{Pairwise Sentence Classification} The training sequence typically consists of two input sentences. Depending on the task these can be:

\begin{itemize}
    \item  In Natural Language Inference (NLI), the two input sentences are the hypothesis and the premise. Depending on the task, the goal can be to determine whether the hypothesis is true (entailment), false (contradiction), or undetermined (neutral) given the premise (MNLI) or whether the hypothesis entails the premise in the form of binary classification (RTE). For these tasks, we generate the OOD data by randomly extracting a sentence from the corpus to serve as the premise and then by random chance construct the hypothesis to either be a slightly changed version of the premise or be a completely new random sentence.
    \item In tasks such as Quora Question Pair (QQP) and Microsoft Research Paraphrase Corpus (MRPC), the goal is to determine if the two input sentences are semantically equivalent or not. We follow a strategy similar to NLI tasks but for the QQP task we post-process the generated sentences by appending a question mark at the end.
\end{itemize}


\textbf{Question NLI}. The goal of this task is to determine if the given paragraph contains the answer to the input question. We sample a paragraph from our corpus and, randomly, either sample a segment from within the paragraph to form a question or sample an unrelated sentence from the corpus. Then, we randomly append a questioning word such as Who, Where, What etc. to the start of the segment and a question mark at the end.

\subsection{Adversarial Training}

\label{sec:adv_train}

Inspired by \cite{micaelli2019zero} and on the promise of adversarial training for NLP \cite{zhu2019freelb}, the key ingredient of our proposed method is to learn a generator that generates training samples. Most methods in adversarial training for NLP~\citep{zhang2020adversarial} perturb the word embeddings instead of generating text due to the discreteness problem of text. In order to generate text, we need an argmax operation which breaks end-to-end differentiability. Since our goal is KD, embedding perturbation introduces the problem of size mismatch between the student and teacher embedding. Instead we generate text and sample from the argmax by using the Gumbel-Softmax distribution \cite{kusner2016gans,jang2016categorical}, a continuous distribution over the simplex that can approximate one-hot samples from a discrete distribution.


\begin{table*}[t]
\centering
\begin{tabular}{lccccc}
\toprule
Task & Model & Method & \makecell{Data\\Generation} & \makecell{Data\\Size} & Score  \\ 
\midrule
\multirow{6}{*}{SST-2}  & Teacher & - & Original & 67K ($\times$1) & 93.0 \\[2pt]
 & Student & - & Original & 67K ($\times$1) & 87.4  \\[2pt]
  & Student & KD & WikiText-103 & 269K ($\times$4) & 84.9  \\[2pt]
  & Student & \makecell{KD + Adv} (Ours) & WikiText-103 & 135K ($\times$2) & 85.0  \\[2pt]
  & Student & \makecell{KD + Adv} (Ours) & WikiText-103 & 269K ($\times$4) & \textbf{85.9}  \\[2pt]
\bottomrule
\end{tabular}
\caption{Results on the single sentence sentiment classification task.}
\label{tab:res1}
\end{table*}

\begin{table*}[t]
\centering
\begin{tabular}{lccccc}
\toprule
Task & Model & Method & \makecell{Data\\Generation} & \makecell{Data\\Size} & Score  \\ 
\midrule
\multirow{6}{*}{MNLI}  & Teacher & - & Original & 392K ($\times$1) & 86.6 \\[2pt]
& Student & - & Original & 392K ($\times$1) & 75.5 \\[2pt]
& Student & KD & WikiText-103 & 1.5M ($\times$4) & 62.5  \\[2pt]
& Student & \makecell{KD + Adv} (Ours) & WikiText-103 & 785K ($\times$2) & 63.8  \\[2pt]
& Student & \makecell{KD + Adv} (Ours) & WikiText-103 & 1.5M ($\times$4) & \textbf{65.1}  \\[2pt]
\midrule
\multirow{6}{*}{RTE}  & Teacher & - & Original & 2.5K ($\times$1) & 70.7 \\[2pt]
& Student & - & Original & 2.5K ($\times$1) & 64.2 \\[2pt]
& Student & KD & WikiText-103 & 10K ($\times$4) & 61.7  \\[2pt]
& Student & \makecell{KD + Adv} (Ours) & WikiText-103 & 5K ($\times$2) & 62.0  \\[2pt]
& Student & \makecell{KD + Adv} (Ours) & WikiText-103 & 10K ($\times$4) & \textbf{62.5}  \\[2pt]
\bottomrule
\end{tabular}
\caption{Results on the NLI classification tasks.}
\label{tab:res2}
\end{table*}

\subsubsection{Generator Pre-training}

Our adversarial generation is closer in spirit to adversarial training and therefore we pre-train the generator to produce samples from the OOD training data $D$~(See Section~\ref{sec:odd}). Specifically, our generator $G$ is a language model which is fed a fixed length sequence of randomly sampled representations from a normal distribution with mean 0 and variance $\sigma^2$ and it generates a sequence of tokens. We train it by minimizing the following loss function:

\begin{equation}
 \mathcal{L}_{PG} \leftarrow \kld{\bo{x}_k}{ \bo{x}_p}
 \label{loss:pre-train}
\end{equation}

where a $x_k$ is a sample from  the OOD training set $D$ and a $x_p$ is a sample from the generator. The algorithm describing this process can be found on Section B of the supplementary material.

\subsubsection{KD with Adversarial Generation}

Once pre-trained, the generator is trained with two losses. The first loss maximises the KL-divergence between the teacher and student model on the generated data. The teacher and student model parameters are fixed. The goal is to generate training samples where the teacher and student diverge the most. However, this can lead to degenerate samples which are not useful for transferring teacher knowledge. The second loss is the same as Equation \ref{loss:pre-train} and prevents the generator from diverging too much from the OOD training data. The overall loss for generator training is thus: 

\begin{align}
\begin{split}
    & \underset{G}{max}\;\; \mathbb{E}_{z_p \sim \mathcal{N}(0,std)}[\kld{T(G(z_p))}{ S(G(z_p))}] \; + \\ & \underset{G}{min}\;\; \mathbb{E}_{z_p \sim \mathcal{N}(0,std)}[\kld{x_k}{G(z_p)}]
\end{split}
\end{align}

where $G$ is the generator, $T$ is the teacher, $S$ is the student, $z_p$ is a sample from normal distribution and $x_k$ is a sample from the OOD training set.

In each training loop we train the generator for $n_G$ steps and the student for $n_S$ steps. Specifically, as shown in Algorithm \ref{alg:distillation} the student is optimized using a joint KD loss between the data samples generated from the generator $G$ and the data samples coming from the OOD dataset. The complete version of our proposed algorithm can be found on Section B of the supplementary material.

\section{Experiments}

We evaluated our proposed adversarial ZSKD approach on six classification tasks from the General Language Understanding Evaluation (GLUE) \citep{wang-etal-2018-glue} benchmark. Specifically, the first task we report results on is binary sentiment classification using the SST-2 \citep{socher-etal-2013-recursive} dataset. In this task, the input is a single sentence and the output is a probability distribution between positive and negative sentiments. 

The second task is ternary natural language inference (NLI) classification on the MNLI \citep{williams-etal-2018-broad} dataset. The input is a pair of sentences and the output is a probability distribution between entailment, contradiction and neutral classes. A similar task is the Recognizing Textual Entailment (RTE) \citep{bentivogli2009fifth} task which is a binary entailment task.

In addition, we report results on the Quora Question Pairs (QQP) \citep{chen-etal:2018:_quora} task and the Microsoft Research Paraphrase Corpus (MRPC) \citep{dolan-brockett-2005-automatically} task which share the goal of predicting semantic equivalence between two sentences. Finally, we report results on the Question Natural Language Inference (QNLI) \citep{wang-etal-2018-glue} dataset which is a binary classification task on predicting whether the answer to a question is found inside a given paragraph. 


\begin{table*}[t]
\centering
\begin{tabular}{lccccc}
\toprule
Task & Model & Method & \makecell{Data\\Generation} & \makecell{Data\\Size} & Score  \\ 
\midrule
\multirow{6}{*}{QQP}  & Teacher & - & Original & 363K ($\times$1) & 89.9 \\[2pt]
& Student & - & Original & 363K ($\times$1) & 83.7 \\[2pt]
& Student & KD & WikiText-103 & 1.4M ($\times$4) & 70.0  \\[2pt]
& Student & \makecell{KD + Adv} (Ours) & WikiText-103 & 728K ($\times$2) & 71.7  \\[2pt]
& Student & \makecell{KD + Adv} (Ours) & WikiText-103 & 1.4M ($\times$4) & \textbf{72.2}  \\[2pt]
\midrule
\multirow{6}{*}{MRPC}  & Teacher & - & Original & 4K ($\times$1) & 87.1 \\[2pt]
& Student & - & Original & 4K ($\times$1) & 78.5 \\[2pt]
& Student & KD & WikiText-103 & 15K ($\times$4) & 74.5  \\[2pt]
& Student & \makecell{KD + Adv} (Ours) & WikiText-103 & 7K ($\times$2) & 75.4  \\[2pt]
& Student & \makecell{KD + Adv} (Ours) & WikiText-103 & 15K ($\times$4) & \textbf{76.4}  \\[2pt]
\bottomrule
\end{tabular}
\caption{Results on the pairwise sentence classification tasks.}
\label{tab:res3}
\end{table*}

\begin{table*}[t]
\centering
\begin{tabular}{lccccc}
\toprule
Task & Model & Method & \makecell{Data\\Generation} & \makecell{Data\\Size} & Score  \\ 
\midrule
\multirow{6}{*}{QNLI}  & Teacher & - & Original & 104K ($\times$1) & 91.5 \\[2pt]
& Student & - & Original & 104K ($\times$1) & 84.1 \\[2pt]
& Student & KD & WikiText-103 & 418K ($\times$4) & 78.1  \\[2pt]
& Student & \makecell{KD + Adv} (Ours) & WikiText-103 & 209K ($\times$2) & 79.1  \\[2pt]
& Student & \makecell{KD + Adv} (Ours) & WikiText-103 & 418K ($\times$4) & \textbf{79.9}  \\[2pt]
\bottomrule
\end{tabular}
\caption{Results on the question NLI task.}
\label{tab:res4}
\end{table*}

\subsection{Experimental Setup}
All models used in this paper are based on two architecture settings from the BERT \citep{devlin-etal-2019-bert} model. Specifically, for the teacher model we used the pre-trained version of the $\text{BERT}_{\text{LARGE}}$ model released by the authors. The model consists of 24 layers. The hidden size is 1024 and the number of heads is 16. The total number of parameters is about 340M. For the student model, we decided to use a significantly smaller version of the BERT model. Specifically, we used the $\text{BERT}_{\text{MINI}}$ version which uses 4 layers with 256 hidden dimension and 4 attention heads. The total size of the model is 11M trainable parameters. Both models use a vocabulary of size 30,522 extracted using the Byte Pair Encoding (BPE) \citep{sennrich-etal-2016-neural} tokenization method.

\paragraph{Hyper-parameters} We fine-tuned the BERT-based student model for 10 epochs and picked the best checkpoint that gave the lowest loss during training. We report results for all methods on the given Dev set. For each task, we selected the best fine-tuning learning rate among 5e-5, 4e-5, 3e-5, and 2e-5 values. We used the AdamW \citep{loshchilov2017decoupled} optimizer with the default values. In addition, we used a linear decay learning rate scheduler with no warmup steps. We set the $\alpha$ values from our algorithm to be 0.2 and the $std$ value to 0.01. Additionally, we set the value $n_G$ to 10 and $n_S$ to 100. Finally, we pre-train the generator for two epochs.

\paragraph{Hardware Details} We trained all models using a single NVIDIA V100 GPU. The batch size was set to 64. We used mixed-precision training \citep{micikevicius2018mixed} to expedite the training procedure. All experiments were run using the PyTorch\footnote{\url{https://pytorch.org/}} framework.

\begin{table*}[t] 
	\centering
    \begin{tabular}{rcccc}
    \toprule
        & neutral & contradiction & entailment & Overall \\ 
        \midrule
        OOD Samples & 67.2  & 68.6 & 51.6 & 62.5 \\ 
        +0.75M Adv & 70.8 & 66.1 & 55.7 & 63.8 \\ 
        +1.5M  Adv & 73.2 & 65.3 & 57.8 & 65.1 \\ 
        \bottomrule
    \end{tabular}
\caption{Per-class F1 scores on MNLI of students trained on OOD samples with incremental subsets of adversarial examples. No Adv refer to the student trained on randomized sentence pairs.}
\label{tab:mnli_per_class}
\end{table*}

\subsection{Results}

Table~\ref{tab:res1} presents our result on SST-2. For all the tasks, we present the original large teacher score, the smaller student score when trained on the training data, the student trained with KD on the OOD data and two experiments with different training set sizes using our algorithm. Our baseline is the KD with OOD data and is adapted from \cite{krishna2019thieves}. Their results were on using same size student and teacher and only on SST-2 and MNLI. We have applied it the smaller student setting and defined the OOD generation process for the 6 other datasets to extend their method.  

The data size (x1, x2 and x4) are the OOD data sizes compared to the task specific training data size. The adversarially trained student, in additional to the OOD data, generates an equal number of adversarial examples. On SST-2, we attain close to the student accuracy using the OOD training data. Our method using x2 OOD data does just as well as the baseline but when we use all the OOD data used by the baseline we increase the accuracy by 1.

The results of the NLI classification tasks, MNLI and RTE, are on Table~\ref{tab:res2}. MNLI is one of the two hardest task that we evaluated on. Looking at the accuracy scores we can see that the student trained on the training data falls well short of the teacher. On this task, we can see the strength of our method as the adversarial training improves the score both when we use x2 OOD data and even further when we use x4 OOD data. High model capacity is important for MNLI. We see a similar trend for RTE.

On pairwise sentence classification, on Table~\ref{tab:res3} we see that MRPC follows a similar trend where the adversarial training algorithm improves the F1 score both when used with x2 OOD data and with x4 OOD data. The same applies for the QQP task. Similar to MNLI, the model capacity and the amount of training data appears to be important for this task. Table~\ref{tab:res4} presents the result on the QNLI task and we see improvements using our algorithm both when using half the OOD data as the baseline and when using the same OOD data. On average we see an improvement of 1.4  over all the tasks.

Overall, we were able to recover between 98.2\% (SST-2) and 86.2\% (MNLI and QQP) of the performance of a version of the student model trained with the original dataset. Similarly, we recovered from 92.3\% (SST-2) to 75.1\% (MNLI) of the performance between the teacher and the final student model.

\subsection{Language Model Generator}

We explored the use of a language model (LM) for OOD generation. Table~\ref{tab:gpt} shows the result of using GPT-2~\cite{radford2019language} as a text generator for SST-2 and MNLI. We do not observe any improvement and the algorithm is much slower in comparison due to the complexity of executing such a large language model. We believe that the reason there is no improvement between OOD data generated from the LM and the data extracted from WikiText-103 is that there is barely any semantic difference between the two OOD generation methods.

\begin{table}[t]
\resizebox{\columnwidth}{!}{%
\centering
\begin{tabular}{l|c|c|c}
Task & Model & LM (GPT-2) & WikiText-103 \\ 
\toprule
SST-2 & Student+KD  & 83.1 & 83.2 \\
MNLI & Student+KD   & 60.0 & 60.2 \\
\end{tabular}}
\caption{Results with using GPT-2 for OOD generation}
\label{tab:gpt}
\end{table}

\subsection{Few-shot Setting}

We explore the strength of our algorithm in the few-shot setting where we have 200 examples per class for both SST-2 and MNLI. Our aim is not to test the limits of our technique by augmenting massive amounts of data but to demonstrate that adversarial training in the few shot setting can improve the accuracy of a model. Table~\ref{tab:abl1} shows the results when we proceed from training on the few samples, to regular KD on the few-samples and finally to our algorithm. Here, we do not do KD on the OOD data, but on the adversarial examples and the few-shot samples. We generate the same number of adversarial examples as the few-shot samples. We can see a consistent improvement for both SST-2 and MNLI.

\begin{table}[t]
\centering
\begin{tabular}{lccc}
\toprule
Task & \makecell{Standard\\Train} & \makecell{KD} & \makecell{KD + Adv} (Ours)  \\
\midrule
SST-2 & 62.0 & 62.8 & \textbf{71.2}  \\
MNLI & 45.9 & 47.0 & \textbf{48.7} \\
\bottomrule
\end{tabular}
\caption{Results on training a student model using 200 examples from each class (few-shot).}
\label{tab:abl1}
\end{table}

\subsection{Analysis}

We inspected the per-class results for MNLI to gain insight into the properties of the adversarially generated samples. Table \ref{tab:mnli_per_class} show that adding adversarial examples continuously improves the performances on \textit{neutral} and \textit{entailment} classes.

Our manual inspection shows that adding the generator to the loop makes the student more robust on examples where the premise and hypothesis doesn't significantly overlap. The gain could be imputable to the diversity of the adversarial examples, although, the generator may produce a nonsensical sequence of words. We observed that the premise and hypothesis rarely share common words, contrary to heuristically populated examples~\footnote{premise and hypothesis are almost identical}. Adversarial examples prevent the student from  relying on the superficial syntactic properties of OOD samples. Section A of the supplementary material presents samples from the adversarial generator for MNLI. 

\section{Conclusion}

We present the first study on Zero-shot Knowledge Distillation (ZSKD) for NLP. We present an algorithm based on OOD data generation and adversarial learning and evaluate on six tasks from the GLUE benchmark reaching to within 75\% of the teacher performance on all tasks while attaining a 30x compression. We believe this is the first step towards ZSKD for NLP-based task. The next steps are to  a) explore a generic methodology for OOD data creation and b) study sequence generation tasks such as Machine Translation and Abstractive Summarization and achieve compression without having access to the original training data.

\bibliographystyle{acl_natbib}
\bibliography{anthology,acl2021}
\newpage
\appendix

\section{Generated Examples}
We present on Table \ref{tab:gen_examples}, a few randomly generated examples using the adversarial generator for the MNLI task. We tested two training settings (1) \textbf{with} and (2) \textbf{without} pre-training the generator and using the $\mathcal{L}_F$ loss during an adversarial step. We can see from the results that without these additional extensions the proposed generator can only generate repeated, frequent tokens. When using these extensions, the generated text is nonsensical, however, it generates diverse words and it is useful for KD. This is reflected on the accuracy of the trained student model which is 56.3\% when trained without pre-training and the $\mathcal{L}_F$ loss. In this scenario, the adversarial step degrades model accuracy. When we train the student model using our proposed algorithm it achieves 65.1\% accuracy.

\begin{table*}[t]
\centering
\begin{tabular}{p{0.45\textwidth}|p{0.45\textwidth}}
\toprule
\multicolumn{2}{c}{\textbf{Pre-training generator and using the $\mathcal{L}_F$ loss during each adversarial step}} \\
\midrule    
\textbf{With} & \textbf{Without}  \\
\midrule
{[CLS] he anime a of survey developedic the headed life raise.,, designingism robot world which 20th the testous out [SEP] swedish ) hudson in studio of pumping pay " a and correspondence " assist,,tion the over and meaningful sponsored. [SEP]} & – – – – – – – – – – – – – – – – – – – – – – – – –ted – – – – – – – – – – – – – – – – – – – – – – – august \\
\midrule
{[CLS] however in and was entertainment service asked of their rebellious def exceptionally carriedlk and to a " in by the quad in. [SEP] after the who bridge which sent direct animal extinct to and april the people bases improvement [PAD] by more his ". [SEP]} & – – – – – – – – – – – – – – – – – – – – – – – – – – – – – – – – – – – – – – – – – – – – – – – – – extended \\
\midrule
{[CLS] so generation included 27 made in the the in of south st johncoef the award his. kin of first (ville [SEP] in to, line – ur wall pt as shy novsky them also of ". helping. a houston 2015 [PAD] [SEP]} & – – – – – – – – – – – – – – – – – – – – – – – – – – – – – – – – – – – – – – – – – – – – – – – – – – \\
\midrule
{[CLS] the expert the victory on - the festival of -, kate like the passry grew must sell in in border longer him [SEP] us king all bb, @ " do [PAD] evaluated,.hai causeway [PAD] ). [PAD] in previousn. [SEP]} & end – – – – – – – – – – – – – – – – – – – – – – – – – – – – – – – – – – – – – – – – – – – – – – – – – \\
\midrule
{[CLS] swiss addition \& moments, from song georgie " analysis the was image in 20th annual fellow as into was of ; pattern. [SEP] cong,, an of provideds her and or. [PAD]leen western classical.. kill the the 2007. [SEP]} & – – – – – – – – – – – – – – – – – – – – – – – – – baronet – – – – – – – – – – – – – – – – – – – – – – – – \\
\midrule
{[CLS] the conducting moderator extracteli, using the war ban civil -oh, then festivals, the golden.ing by positively. [SEP] besides labour oak 600 patton later from. the analysis [PAD] finish barking base. mark [PAD] his $\times$ 23 s ). [SEP]} & – – – – – – – – – – – – – – – – – – – – – – – – – – – – – – – – – – – – – – – – – – – – – – – – – paid \\
\bottomrule
\end{tabular}
\caption{Generated examples using the adversarial generator for MNLI.}
\label{tab:gen_examples}
\end{table*}

\section{Algorithm}

In this section, we present the complete algorithm of our proposed Zero-shot Knowledge Distillation process. We include the process of pre-training the generator along the adversarial train used for improving the student performance. The complete procedure can be seen on Algorithm \ref{alg:distillation_complete}.

\begin{algorithm*}[t]
  \SetAlgoLined
  \SetKwInput{pretrain}{pretrain}
  \SetKwInput{data}{dataset}
  \SetKwInput{init}{initialize}
  \pretrain{$T(\cdot)$}
  \data{$D$}
  \init{$G(\cdot; \phi)$}
  \init{ $S(\cdot; \theta)$}
  \BlankLine
  \BlankLine
  \# Pre-train Generator \;
  \For{$k \leftarrow 1, 2, ..., N$}{
  $\bo{z} \leftarrow \{\bo{z_0,\ldots,z_l}\} \sim \mathcal{N}(\bo{0},\bo{std})$ \;
   $\bo{x}_k \leftarrow D$ \;
   $\bo{x}_p \leftarrow G(\bo{z}; \phi)$ \;
   $\mathcal{L}_{PG} \leftarrow \kld{\bo{x}_k}{ \bo{x}_p}$ \;
    $\phi \leftarrow \phi - \lambda \dfrac{\partial \mathcal{L}_{PG}}{\partial \phi}$ \;
    decay $\lambda$ \;
  }
  \BlankLine
  \BlankLine
  \# Adversarial Train \;
  \For{$k \leftarrow 1, 2, ..., N$}{
    $\bo{x}_k \leftarrow D$ \;
    \BlankLine
    \For{$1, 2, ..., n_G$}{
        $\bo{z} \leftarrow \{\bo{z_0,\ldots,z_l}\} \sim \mathcal{N}(\bo{0},\bo{std})$ \;
        $\bo{x}_\text{logits} \leftarrow G(\bo{z}; \phi)$ \;
        $\bo{x}_p \leftarrow \text{Gumbel-Softmax}(\bo{x}_\text{logits})$ \;
        $\mathcal{L}_A \leftarrow - \kld{T(\bo{x}_p)}{ S(\bo{x}_p)}$ \;
        $\mathcal{L}_F \leftarrow \kld{\bo{x}_k}{ \bo{x}_p}$ \;
        $\mathcal{L}_T \leftarrow \frac{\mathcal{L}_A + \mathcal{L}_F}{2} $ \;
        $\phi \leftarrow \phi - \eta \dfrac{\partial \mathcal{L}_T}{\partial \phi}$ \;
    }
    \BlankLine
    \BlankLine
    \For{$1, 2, ..., n_S$}{
        $\bo{z} \leftarrow \{\bo{z_0,\ldots,z_l}\} \sim \mathcal{N}(\bo{0},\bo{std})$ \;
        $\bo{x}_\text{logits} \leftarrow G(\bo{z}; \phi)$ \;
        $\bo{x}_p \leftarrow \text{Gumbel-Softmax}(\bo{x}_\text{logits})$ \;
        \BlankLine
        $\mathcal{L}_G \leftarrow \kld{T(\bo{x}_p)}{ S(\bo{x}_p)}$ \;
        $\mathcal{L}_{OOD} \leftarrow \kld{T(\bo{x}_k)}{ S(\bo{x}_k)}$ \;
        $\mathcal{L} \leftarrow \alpha \cdot \mathcal{L}_G + (1-\alpha) \cdot \mathcal{L}_{OOD}$ \;
        $\theta \leftarrow \theta - \eta \dfrac{\partial \mathcal{L}}{\partial \theta}$ \;
    }
    \BlankLine
    decay $\eta$ \;
    \BlankLine
    \BlankLine
 }
 \caption{Zero-shot KD (Complete)}
 \label{alg:distillation_complete}
\end{algorithm*}

\end{document}